\documentclass[a4paper]{article}

\usepackage{authblk}
\usepackage[utf8]{inputenc}
\usepackage[T1]{fontenc}
\usepackage{pslatex}
\usepackage[english]{babel}
\usepackage{fancyhdr}
\usepackage{hyperref}
\usepackage{amsmath,amssymb,amsfonts}
\usepackage{graphicx}
\usepackage{textcomp}
\usepackage{xcolor}
\usepackage{algorithm}
\usepackage[noend]{algpseudocode}
\usepackage{url}
\usepackage{hyperref}
\usepackage{microtype}
\usepackage{subcaption}
\usepackage{tikz}
\usetikzlibrary{positioning}
\usetikzlibrary{decorations.pathreplacing}
\usepackage{cite}
\usepackage{wasysym}
\usepackage{booktabs}
\usepackage{paralist}
\usepackage{hyperref}

\newcommand{\N}{\mathbb{N}}
\newcommand{\F}{\mathbb{F}}

\newtheorem{problem}{Problem}
\newtheorem{proposition}{Proposition}

\providecommand{\keywords}[1]{\textbf{\textit{Keywords }} #1}

\begin{document}

\title{Evolutionary Strategies for the Design of Binary Linear Codes}

\author[1,2]{Claude Carlet}
\author[3]{Luca Mariot}
\author[4]{Luca Manzoni}
\author[3]{Stjepan Picek}

\affil[1]{{\normalsize Department of Mathematics, Universit\'{e} Paris 8, 2 rue de la libert\'{e}, 93526 Saint-Denis Cedex, France}}

\affil[2]{{\normalsize University of Bergen, Bergen, Norway} \\
	
	{\small \texttt{claude.carlet@gmail.com}}}

\affil[3]{{\normalsize Digital Security Group, Radboud University, Postbus 9010, 6500 GL Nijmegen, The Netherlands} \\
	
	{\small \texttt{\{luca.mariot,stjepan.picek\}@ru.nl}}}

\affil[4]{{\normalsize Dipartimento di Matematica e Geoscienze, Universit\`{a} degli Studi di Trieste, Via Valerio 12/1, 34127 Trieste, Italy} \\
	
	{\small \texttt{lmanzoni@units.it}}}
	
\maketitle

\begin{abstract}
\noindent
The design of binary error-correcting codes is a challenging optimization problem with several applications in telecommunications and storage, which has also been addressed with metaheuristic techniques and evolutionary algorithms. Still, all these efforts focused on optimizing the minimum distance of unrestricted binary codes, i.e., with no constraints on their linearity, which is a desirable property for efficient implementations. In this paper, we present an Evolutionary Strategy (ES) algorithm that explores only the subset of linear codes of a fixed length and dimension. To that end, we represent the candidate solutions as binary matrices and devise variation operators that preserve their ranks. Our experiments show that up to length $n=14$, our ES always converges to an optimal solution with a full success rate, and the evolved codes are all inequivalent to the Best-Known Linear Code (BKLC) given by MAGMA. On the other hand, for larger lengths, both the success rate of the ES as well as the diversity of the evolved codes start to drop, with the extreme case of $(16,8,5)$ codes which all turn out to be equivalent to MAGMA's BKLC.
\end{abstract}


\section{Introduction}
\label{sec:intro}

A central problem in information theory is the transmission of messages over noisy channels. To this end, error-correcting codes are usually employed to add redundancy to a message before sending it over a channel. A common setting is to consider messages over the binary alphabet $\F_2=\{0,1\}$, under the hypothesis of a binary symmetric channel~\cite{Mceliece04}. To be useful in practice, a binary code must have the following properties: (a) a high minimum Hamming distance, (b) a high number of codewords, and (c) an efficient decoding algorithm. While (a) and (b) induce a direct trade-off---intuitively, the more codewords belong to a code, the more closely packed they must be---property (c) is usually addressed by requiring that the code is linear, i.e., that it forms a $k$-dimensional subspace of $\F_2^n$.

The design of a good binary code can be seen as a combinatorial optimization problem, where the objective is to maximize the minimum distance of a set of codewords. This problem is known to be $\mathcal{NP}$-hard since it is equivalent to finding a maximum clique in a graph~\cite{Karp72}. To date, several optimization algorithms have been applied to solve this problem, including evolutionary algorithms~\cite{DontasJ90,McGuireS98,ChenFW98,McCarneyHR12} and other metaheuristics~\cite{GamalHSW87,ChenFW98,AlbaC04,Cotta04,BlumBR05}. To the best of our knowledge, most of these works target the construction of unrestricted binary codes without enforcing any requirement on the linearity of the resulting code. The only exception that we are aware of is~\cite{McGuireS98}, where genetic algorithms are used to evolve the generator matrices of linear codes. However, this approach does not preserve the rank of the matrices, and thus the dimension of the evolved codes can vary.

In this paper, we propose for the first time an Evolutionary Strategy (ES) algorithm for the design of binary linear codes with the best possible minimum distance $d$ for a given combination of length $n$ and dimension $k$. We adopt a combinatorial representation that allows the ES to explore only the set of $k$-dimensional subspaces of $\F_2^n$. The ES encodes a candidate solution as a $k\times n$ binary generator matrix of full rank $k$. On account of a recent result proved in~\cite{Carlet22}, the fitness of a matrix is evaluated as the number of monomials of degree less than $d$ in the Algebraic Normal Form (ANF) of the indicator Boolean function of the linear code. Next, for the variation operators, we consider both a classical ES using only mutation and a variant combining mutation and crossover. Both types of operators are designed so that the rank of the offspring matrices is preserved.

We evaluate experimentally our approach over five different instances of linear codes for lengths ranging from $n=12$ to $n=16$ and dimension set to $k = \lfloor \frac{n}{2} \rfloor$, seeking to reach the bounds on the minimum distance reported in~\cite{Grassl07}. We test four different versions of our ES algorithm, depending on the replacement strategy (either $(\mu, \lambda)$ or $(\mu+\lambda)$) and whether the crossover is applied or not. All variants achieve a full success rate up to length $n=14$. Somewhat surprisingly, for the larger instances $n=15$ and $n=16$, the simple $(\mu,\lambda)$ without crossover scores the best performance. Indeed, we observe that the average fitness and distance of the population in the $(\mu+\lambda)$ variants quickly converge to a highly fit, low-diversity area of the search space.

Finally, we investigate the diversity of the codes evolved by our ES algorithm up to isomorphism. In particular, we test how many of our codes are inequivalent to the Best-Known Linear Code (BKLC) constructed through the MAGMA computer algebra system, and we group them into equivalence classes. Interestingly, for lengths up to $n=14$, all codes turn out to be inequivalent to MAGMA's BKLC, and they belong to a high number of equivalence classes. The situation is, however, reversed for the larger instances: while for $n=15$, there is still a good proportion of inequivalent codes, for $n=16$, all codes are equivalent to MAGMA's BKLC. Therefore, our ES essentially converges to the same solution.

\section{Background}
\label{sec:bg}

In this section, we cover all background information related to binary linear codes and Boolean functions that we will use in the paper. The treatment is far from exhaustive, and we refer the reader to~\cite{Mceliece04,HuffmanP10} and~\cite{Carlet21} for a more complete survey of the main results on error-correcting codes and Boolean functions, respectively.

\subsection{Binary Linear Codes}
\label{subsec:lin-codes}
Let $\F_2 = \{0,1\}$ denote the finite field with two elements. For any $n \in \N$, the $n$-dimensional vector space over $\F_2$ is denoted by $\F_2^n$, where the sum of two vectors $x, y \in \F_2^n$ corresponds to their bitwise XOR, while the multiplication of $x$ by a scalar $a \in \F_2$ is defined as the logical AND of $a$ with each coordinate of $x$. The Hamming distance $d_H(x,y)$ of two vectors $x, y \in \F_2^n$ is the number of coordinates where $x$ and $y$ disagree. Given $x \in \F_2^n$, the Hamming weight of $x$ is the number of its nonzero coordinates, or equivalently the Hamming distance $d_H(x, \underline{0})$ between $x$ and the null vector $\underline{0}$.

A binary code of length $n$ is any subset $C$ of $\F_2^n$. The elements of $C$ are also called codewords, and the size of $C$ is usually denoted by $M$. The minimum distance $d$ of $C$ is the minimum Hamming distance between any two codewords $x,y \in C$. One of the main problems in coding theory is to determine what is the maximum number of codewords $A(n,d)$ that a code $C$ of length $n$ and minimum distance $d$ can have. Several theoretical bounds exist on $A(n,d)$. For example, the Gilbert-Varshamov bound, and the Singleton bound, respectively, give a lower and an upper bound on $A(n,d)$ as follows:
\begin{equation}
	\label{eq:gvs-bounds}
	\frac{2^n}{\sum_{i=0}^{d-1} \binom{n}{i}} \le A(n,d) \le 2^{n-d+1} \enspace .
\end{equation}
More refined bounds exist, such as the Hamming bound, for which we refer the reader to~\cite{HuffmanP10}. A slightly different but equivalent problem is to fix the length $n$ and size $M$ of a code and then maximize the resulting minimum distance $d$ according to an analogous upper bound.

A binary code $C \subseteq \F_2^n$ is called linear if it forms a vector subspace of $\F_2^n$. In this case, the size of the code can be compactly described by the dimension $k \le n$ of the subspace. Indeed, the encoding process amounts to multiplying a $k$-bit vector $m \in \F_2^k$ by a $k\times n$ generator matrix $G_C$ which spans the code $C$ (and, therefore, $G_c$ has rank $k$). The resulting $n$-bit vector $c = mG_C \in C$ will be the codeword corresponding to the message $m$. Thus, the size of $C$ is $M=2^k$. A linear code of length $n$, dimension $k$, and minimum distance $d$ is also denoted as an $(n,k,d)$ code. Some of the bounds mentioned above can be simplified if one is dealing with a linear code. For example, the Singleton bound for a $(n,k,d)$ linear code becomes
\begin{equation}
	\label{eq:sing}
	k \le n-d+1 \enspace ,
\end{equation}
which also gives an upper bound on the minimum distance $d$ as $d \le n-k+1$. 

\subsection{Boolean Functions}
\label{subsec:bool-fun}
A Boolean function of $n \in \N$ variables is a mapping $f: \F_2^n \to \F_2$, i.e., a function associating to each $n$-bit input vector a single output bit. The support of $f$ is defined as $supp(f) = \{x \in \F_2^n: f(x) \neq 0\}$, i.e., the set of all input vectors that map to $1$ under $f$. The most common way to represent a Boolean function is by means of its truth table: assuming that a total order is fixed on $\F_2^n$ (e.g., the lexicographic order), then the truth table of $f$ is the $2^n$-bit vector specifying for each input vector $x \in \F_2^n$ the corresponding output value $f(x) \in \F_2$. 

A second representation method for Boolean functions commonly used in cryptography and coding theory is the Algebraic Normal Form (ANF). Given $f: \F_2^n \to \F_2$, the ANF of $f$ is the multivariate polynomial in the quotient ring $\F_2[x_1,\ldots,x_n]/(x_1\oplus x_1^2, \ldots, x_n \oplus x_n^2)$ defined as follows:
\begin{equation}
	P_f(x) = \bigoplus_{I \in 2^{[n]}} a_I \left( \prod_{i \in I} x_i \right) \enspace ,
\end{equation}
with $2^{[n]}$ being the power set of $[n] = \{1,\cdots,n\}$. The coefficients $a_I \in \F_2^n$ of the ANF can be computed from the truth table of $f$ via the M\"{obius} transform:
\begin{equation}
	\label{eq:mobius}
	a_I = \bigoplus_{x \in \F_2^n: supp(x) \subseteq I} f(x) \enspace ,
\end{equation}
where $supp(x) = \{i \in [n]: x_i \neq 0 \}$ denotes the support of $x$, or equivalently the set of nonzero coordinates of $x$.

The degree of a coefficient $a_I$ corresponds to the size of $I$ (that is, to the number of variables in the corresponding monomials). Then, the algebraic degree of $f$ is defined as the largest monomial occurring in the ANF of $f$, or equivalently as the cardinality of the largest $I \in 2^{[n]}$ such that $a_I \neq  0$.

The algebraic normal form of Boolean functions can be used to characterize the minimum distance of binary linear codes, as shown by C. Carlet~\cite{Carlet22}:
\begin{proposition}
	\label{prop:carlet}
	Let $C \subseteq \F_2^n$ be a $(n,k,d)$ binary linear code, and define the indicator of $C$ as the Boolean function $1_C: \F_2^n \to \F_2$ whose support coincides with the code, i.e., $supp(1_C) = C$. Then,
	\begin{equation}
		\label{eq:mindist}
		d = \min \{ |I| \in 2^{[n]} : a_I = 0\} \enspace ,
	\end{equation}
	where $a_I$ denotes the coefficients of the ANF of $1_C$.
\end{proposition}
In other words, one can check if a binary linear code of length $n$ and dimension $k$ has minimum distance $d$ by verifying that all monomials of degree strictly less than $d$ appear in the ANF of the indicator function $1_C$, while the smallest non-occurring monomial has degree $d$. This observation will be used in the next sections to define a fitness function for our optimization problem of interest.

\section{Related Works}
\label{sec:rel-works}

El Gamal et al.~\cite{GamalHSW87} were the first to investigate the design of unrestricted codes by simulated annealing (SA). Their results showed that SA was capable of finding many new constant-weight and spherical codes, in some cases improving on the known bounds for $A(n,d)$. The first application of Genetic Algorithms (GAs) to evolve error-correcting codes with maximal distance was proposed by Dontas and De Jong~\cite{DontasJ90}. The authors encoded a candidate solution as a bitstring of length $n\cdot M$, representing the concatenation of $M$ codewords of length $n$, and maximized two fitness functions based on the pairwise Hamming distance between codewords. Using the same encoding and fitness, Chen et al.~\cite{ChenFW98} developed a hybrid algorithm combining GA and SA to design error-correcting codes.

McGuire and Sabin~\cite{McGuireS98} employed a GA to search for linear codes. To enforce the linearity of the evolved codes, the genotype of the candidate solutions were $k\cdot n$ bitstrings, which represented the concatenation of the rows of $k \times n$ generator matrices. However, the authors used crossover and mutation operators that do not preserve the ranks of the resulting matrices. Therefore, the dimension of the codes evolved by their GA is not fixed, contrary to what is claimed in the paper.

Alba and Chicano~\cite{AlbaC04} designed a so-called Repulsion Algorithm (RA) for the error-correcting codes problem that takes inspiration from electrostatic phenomena. In particular, the codewords are represented as particles obeying Coulomb law, and their next position on the Hamming cube is determined by computing the resultant force vectors. 
Cotta~\cite{Cotta04} experimented with several combinations of Scatter Search (SS) and Memetic Algorithms (MAs) to evolve codes of length up to $n=20$ and $M=40$ codewords. The results indicated that both SS and MAs could outperform other metaheuristics on this problem.

Blum et al.~\cite{BlumBR05} investigated an Iterated Local Search (ILS) technique and combined it with a constructive heuristic to design error-correcting codes, showing that it achieved state-of-the-art performances.
McCarney et al.~\cite{McCarneyHR12} considered GAs and Genetic Programming (GP) to evolve binary codes, where the chromosome's genes represent the entire codeword rather than a single symbol, as in most other approaches. The results on codes of length $n=12,13,$ and $17$ and minimum distance $6$ suggested that GP can outperform GA on this problem.  

A few works address the construction of combinatorial designs that are analogous to error-correcting codes through evolutionary algorithms. For example, Mariot et al.~\cite{MariotPJL18} employed GA and GP to evolve binary orthogonal arrays, which are equivalent to binary codes. Knezevic et al.~\cite{KnezevicPMJL18} considered using Estimation of Distribution Algorithms (EDAs) to design disjunct matrices, which can be seen as superimposed codes. More recently, Mariot et al.~\cite{MariotPJDL22} proposed an evolutionary algorithm for the incremental construction of a permutation code, where the codewords are permutations instead of binary vectors.

Although in this paper we use Boolean functions to compute the fitness, we note that their construction via metaheuristics is also a solid research thread, especially concerning the optimization of their cryptographic properties. We refer the reader to~\cite{MariotJBH22} for a survey of the main results in this area.

\section{Evolutionary Strategy Algorithm}
\label{sec:es}

In this section, we describe the main components of our Evolutionary Strategy (ES) algorithm used to evolve binary linear codes. As a reference for later, we formally state the optimization problem that we are interested in as follows:
\begin{problem}
	\label{pb:stat}
	Let $n,k \in \N$ with $k\le n$. Find a $(n,k,d)$ binary linear code $C \subseteq \F_2^n$ reaching the highest possible minimum distance $d$.
\end{problem}
Remark that the upper and lower bounds on the highest minimum distance mentioned in Section~\ref{subsec:lin-codes} are not tight in general. However, for binary codes of length up to $n=256$, one can use the tables provided by Grassl~\cite{Grassl07} to determine the best-known values. In particular, for the combinations of $n$ and $k$ that we will consider in our experiments in Section~\ref{sec:exp}, the lower and upper bounds on $d$ coincide. So, Problem~\ref{pb:stat} is well-defined for all instances considered in our tests. 

\subsection{Solutions Encoding and Search Space}
\label{subsec:enc}
As we discussed in Section~\ref{sec:rel-works}, most of the works addressing the design of error-correcting codes via metaheuristic algorithms usually target generic codes without any constraint on their linearity. The only exception seems to be the paper by McGuire and Sabin~\cite{McGuireS98} where a GA evolves a generator matrix, but there is no control on the dimension $k$ of the corresponding code. As a matter of fact, if one applies unrestricted variation operators on a matrix of rank $k$ (such as one-point crossover or bit-flip mutation), then the vectors in the resulting matrix might not be linearly independent, and thus the associated code could have a lower dimension. This might, in turn, pose issues because the optimal value of the fitness function (which is related to the best-known minimum distance for a given combination of $n$ and $k$) can change during the evolution process.

In our approach, we consider the genotype of a candidate solution $S$ as a $k\times n$ binary matrix $G$ of full rank $k$. Accordingly, the phenotype corresponding to $G$ is the code $C$ which is the image of the linear map defined by $G$. Formally, we can define the phenotype code as:
\begin{equation}
	C = \{c \in \F_2^n: c = x \cdot G, \ x \in \F_2^k\} \enspace .
\end{equation}
Therefore, the search space $\mathcal{S}_{n,k}$ for a given combination of code length $n$ and dimension $k$ is the set of all $k\times n$ binary matrices of rank $k$, or equivalently the set of all $k$-dimensional subspaces of $\F_2^n$, also called the Grassmannian $Gr(\F_2^n,k)$ of $\F_2^n$. It is known that the size of this space equals the Gaussian binomial coefficient $\binom{n}{k}_2$, defined as~\cite{MullenP2013}:
\begin{equation}
	\label{eq:gauss-bc}
	\binom{n}{k}_2 = \frac{(2^n-1)(2^{n-1}-1)\cdots(2^{n-k+1}-1)}{(2^k-1)(2^{k-1}-1)\cdots(2^{k-(k-2)}-1)} \enspace .
\end{equation}
It is clear from the expression above that the size of the Grassmannian grows very quickly, and thus exhaustive search in this space becomes unfeasible already for small values of $n$ and $k$. For example, setting $n=12$ and $k=6$, the corresponding Grassmannian is composed of about $2.3 \cdot 10^{11}$ (230 billion) subspaces. This observation provides a basic argument motivating the use of heuristic optimization algorithms to solve Problem~\ref{pb:stat}. 

In what follows, we will endow the Grassmannian $Gr(\F_2^n,k)$ with a distance, turning it into a metric space. This will be useful to study the diversity of the population evolved by the ES. In particular, let $A,B \subseteq \F_2^n$ be two $k$-dimensional subspaces of $\F_2^n$. Then, the distance between $A$ and $B$ equals:
\begin{equation}
	\label{eq:dist-sub}
	d(A,B) = dim(A) + dim(B) - 2dim(A \cap B) = 2(k - dim(A \cap B)) \enspace .
\end{equation}
This distance was introduced by K\"{o}tter and Kschischang in~\cite{KoetterK07} to study error-correction in the setting of random network linear coding, where the transmitted codewords are not vectors of symbols, but rather vector subspaces themselves. Hence, the problem becomes to find a set of subspaces in the projective space of $\F_2^n$ (that is, the set of all subspaces of $\F_2^n$), which are far apart from each other under the distance defined in Eq.~\eqref{eq:dist-sub}. In particular, this distance is the length of a geodesic path joining the two subspaces $A$ and $B$ when seen as points on the poset (partially ordered set) of the projective space of $\F_2^n$, where the elements are ordered with respect to subset inclusion. In our case, the search space $\mathcal{S}_{n,k} = Gr(\F_2^n, k)$ corresponds to the antichain of this poset that includes all $k$-dimensional subspaces of $\F_2^n$. 

\subsection{Fitness Function}
\label{subsec:fitness}
Many of the works surveyed in Section~\ref{sec:rel-works} optimize binary codes by maximizing a fitness function that directly measures the pairwise Hamming distance between the codewords. In this work, on the other hand, we define a new fitness function that is based on the characterization of the minimum distance in terms of the ANF of the indicator function proved in~\cite{Carlet22}. In particular, we use the fitness function to count the number of coefficients of degree strictly less than $d$ that occur in the ANF of the indicator, with the objective of maximizing it. Formally, given a code $C$ of length $n$ and dimension $k$, and denoting by $a_I$ the monomial in the ANF of the indicator function $1_C$ for $I \in 2^{[n]}$, the fitness of $C$ is defined as:
\begin{equation}
	\label{eq:fit}
	fit(C) = \{I \in 2^{[n]}: |I| < d , \ a_I \neq 0\}| \enspace .
\end{equation}
By Proposition~\ref{prop:carlet}, $C$ is a $(n,k,d)$ linear code if and only if all coefficients of degree less than $d$ are in the ANF of its indicator, and the optimal value for $fit$ is:
\begin{equation}
	\label{eq:fit-opt}
	fit^*_{n,d} = \sum_{i=0}^{d-1} \binom{n}{i} \enspace .
\end{equation}

To summarize, the fitness value of a solution $G \in \mathcal{S}_{n,k}$ is computed as follows:
\begin{compactenum}
	\item Generate the linear code $C$ as the subspace spanned by the matrix $G$.
	\item Write the truth table of the indicator function $1_C: \F_2^n \to \F_2$ by setting $f(x) = 1$ if $x \in C$, and zero otherwise.
	\item Compute the ANF of $1_C$ using the Fast M\"{o}bius Transform algorithm.
	\item Compute $fit(C)$ using Eq.~\eqref{eq:fit}.
\end{compactenum}
Although the Fast M\"{obius} Transform yields a significant improvement in the time complexity required to compute the ANF over the naive procedure, it is still computationally cumbersome to compute it for Boolean functions with a relatively high number of variables. For this reason, in our experiments, we limit ourselves to linear codes of length up to $n=16$.

\subsection{Rank-Preserving Mutation and Crossover}
\label{subsec:var-op}
We now describe the variation operators that we employed to generate new candidate solutions from a population of $k$-dimensional subspaces of $\F_2^n$, represented by their generator matrices.

For mutation, we adopt the same operator proposed in~\cite{MariotSLM22}: there, the authors were interested in evolving an invertible binary matrix that was used to define an affine transformation for a bent Boolean function. The operator can be straightforwardly adapted to our problem, even though we are not dealing with invertible matrices. Indeed, the basic principle of~\cite{MariotSLM22} is to preserve the invertibility of a square $n\times n$ binary matrix by keeping its rank equal to $n$. In our case, we use the same idea to maintain the rank of a rectangular matrix. Specifically, given a $k\times n$ binary matrix $G$ of full rank $k$ and a row $i \in [k]$, the mutation operator samples a random number $r \in (0,1)$ and checks whether it is smaller than the mutation probability $p_{mut}$. If this is the case, $G$ is mutated as follows:
\begin{compactenum}
\item Remove the $i$-th row of $G$, obtaining a $(k-1)\times n$ matrix $G'$ of rank $k-1$.
\item Generate the subspace spanned by $G'$, denoted as $span(G')$, and compute the complement $\mathcal{C} = \F_2^n \setminus span(G')$.
\item Pick a random vector $v \in \mathcal{C}$ and insert it in $G'$ as a row in position $i$, obtaining the mutated $k\times n$ matrix $H$.
\end{compactenum}
By construction, the random vector $v$ sampled in step 3 is linearly independent with all vectors in the span of $G'$. Therefore, the mutated matrix $H$ has the same rank $k$ as the original matrix $G$. The process is then repeated for all rows $i \in [k]$. One can notice that the two matrices $G$ and $H$ are at distance $2$ under the metric of Eq.~\eqref{eq:dist-sub}, which is the minimum possible for distinct points in the Grassmannian $Gr(\F_2^n, k)$~\cite{KoetterK07}. Therefore, this operator effectively perturbs a candidate solution by transforming it into one of its closest neighbors.

Given two $k\times n$ parent matrices $G_1, G_2$ of rank $k$, our crossover operator generates an offspring matrix $H$ in the following way:
\begin{compactenum}
\item Concatenate the rows of $G_1$ and $G_2$, thus obtaining a $2k \times n$ matrix $J$.
\item Perform a random shuffle of the rows in $J$.
\item Generate $H$ by selecting a subset of $k$ linearly independent vectors from $J$.
\end{compactenum}
Step 3 is performed incrementally: the offspring matrix $H$ is filled by adding the rows of $J$ from top to bottom, checking if the current row is linearly independent with all previously added ones. If it is not, then the row is discarded, and the next one is attempted. Notice that it is always possible to find a set of $k$ linearly independent vectors in $J$ to construct $H$ since both $G_1$ and $G_2$ have rank $k$. Thus, the worst case arises when all rows of one of the parents are in the span of the other (i.e., $G_1$ and $G_2$ generate the same subspace). In this situation, the offspring will also end up spanning the same subspace, although the generator matrix might look different from the parents. From a linear algebraic point of view, this eventuality corresponds to a change of basis on the same subspace.
By the above argument, it follows that also the crossover operator preserves the rank $k$ of the parents in the offspring.

\section{Experiments}
\label{sec:exp}

\subsection{Experimental Setting}
\label{subsec:exp-set}
Evolutionary strategies are specified by two main parameters, the population size $\lambda$ and the reproduction pool size $\mu$~\cite{Luke15}. At each generation, the $\mu$ best parents in the population are selected for reproduction (truncation selection). Then, in the $(\mu,\lambda)$-ES variant each selected individual generates $\mu/\lambda$ offspring individuals, and their fitness is evaluated. In this case, the new offspring entirely replaces the old population, and the process is then iterated. The $(\mu+\lambda)$-ES variant differs from the fact that the $\mu$ parents from the old population are brought into the new population. To keep the population size fixed to $\lambda$ in the $(\mu+\lambda)$-ES variant in our experiments, we generate $(\lambda/\mu) - 1$ offspring individuals for each selected parent. In classical ES, the parents usually create the offspring only by applying a mutation operator. We also considered a variant of ES augmented with crossover, which works as follows: each parent generates an offspring matrix by first performing crossover with a random mate selected from the reproduction pool of the $\mu$ best individuals. Then, mutation is applied as usual. Therefore, in our experiments we considered four variants of ES, depending on the replacement mechanism ($(\mu, \lambda)$ or $(\mu+\lambda))$, and whether crossover $(\chi)$ is applied or not.

Concerning the combinations of length $n$, dimension $k$, and minimum distance $d$ of the codes, we experimented over five problem instances: $(12, 6, 4)$, $(13, 6, 4)$, $(14, 7, 4)$, $(15, 7, 5)$, and $(16, 8, 5)$. In particular, we always set $k = \lfloor n/2 \rfloor$ since this gives the largest search space possible for a given $n$. Starting from $n=12$ yields the smallest instance that is not amenable to exhaustive search. The corresponding minimum distance $d$ (that represents the optimization objective) has been taken from the tables reported in~\cite{Grassl07}. In all these cases, the lower and upper bounds on $d$ coincide, so these are the best minimum distances one can get for these combinations of $n$ and $k$. For the ES parameters, we set the population size $\lambda$ equal to the length $n$, and $\mu = \lfloor n/3 \rfloor$. The mutation probability was set to $p_{mut} = 1/n$. These are the same parameters settings adopted for the ES in~\cite{MariotSLM22} to evolve invertible binary matrices, and after a preliminary tuning phase we noticed that they also worked well on Problem~\ref{pb:stat}. The fitness budget was set to $20\, 000$ generations of the ES algorithm, since we remarked that the best fitness seldom improved after this threshold. Finally, we repeated each experiment over $100$ independent runs to get statistically sound results.

Table~\ref{tab:param} summarizes our experimental design with all relevant parameters, along with the size of the corresponding search space $\mathcal{S}_{n,k}$ and the value of the best fitness value for each considered problem instance.

\begin{table}
\centering
\scriptsize
\label{tab:param}
\begin{tabular}{cccccc}
\hline
$(n,k,d)$ & $\# \mathcal{S}_{n,k}$ & $fit^*_{n,d}$ & $\lambda$ & $\mu$ & $p_{mut}$ \\
\hline
$(12, 6, 4)$ & $2.31 \cdot 10^{11}$ & 299 & $12$ & $4$ & 0.083 \\
$(13, 6, 4)$ & $1.49 \cdot 10^{13}$ & 378 & $13$ & $4$ & 0.077 \\
$(14, 7, 4)$ & $1.92 \cdot 10^{15}$ & 470 & $14$ & $4$ & 0.071 \\
$(15, 7, 5)$ & $2.47 \cdot 10^{17}$ & 1941 & $15$ & $5$ & 0.067 \\
$(16, 8, 5)$ & $6.34 \cdot 10^{19}$ & 2517 & $16$ & $5$ & 0.063 \\
\hline
\end{tabular}
\caption{Summary of the parameter settings, search space size, and best fitness value for each problem instance.}
\end{table}

\subsection{Results}
\label{subsec:res}
Table~\ref{tab:rates} reports the success rates of the four ES variants over 100 independent runs for the five considered problem instances, that is, how many times they converged to an optimal linear code. We denote a crossover-augmented ES variant by appending $+\chi$ to it.

\begin{table}
	\centering
    \scriptsize
	\begin{tabular}{ccccc}
		\hline
		$(n,k,d)$ & $(\mu, \lambda)$-ES & $(\mu, \lambda)$+$\chi$-ES & $(\mu+\lambda)$-ES & $(\mu+\lambda)$+$\chi$-ES \\
		\hline
		$(12,6,4)$ & 100 & 100 & 100 & 100 \\
		$(13,6,4)$ & 100 & 100 & 100 & 100 \\
		$(14,7,4)$ & 100 & 100 & 100 & 100 \\
		$(15,7,5)$ & 100 & 100 & 77  & 81 \\
		$(16,8,5)$ & 92  & 76  & 18  & 17 \\
		\hline
	\end{tabular}
	\caption{Success rates (over 100 runs) of the four considered ES variants.}
	\label{tab:rates}
\end{table}

The first interesting remark is that all ES variants always converge to an optimal solution up to length $n=14$, seemingly indicating that Problem~\ref{pb:stat} is rather easy on these problem instances, independently of the replacement mechanism and the use of crossover. For $(15,7,5)$, the $(\mu,\lambda)$ variants still converge in all runs, while the $(\mu+\lambda)$-ES achieve a lower success rate, although still quite high. The biggest difference can be seen on the largest problem instance $(16,8,5)$. In this case, the only variant reaching a very high success rate of $92\%$ is the $(\mu,\lambda)$-ES. Somewhat surprisingly, adding crossover to this variant actually worsens the performance. On  the other hand, the $(\mu+\lambda)$-ES variants reach a very low success rate on this instance, independently of crossover. Therefore, in general the main factor influencing the performance is the replacement mechanism, rather than crossover. Apparently, letting the parents directly compete with their children as in the $(\mu+\lambda)$ variant is detrimental for this particular optimization problem.

\begin{figure}[t]
	\centering
	\includegraphics[width=0.9\textwidth]{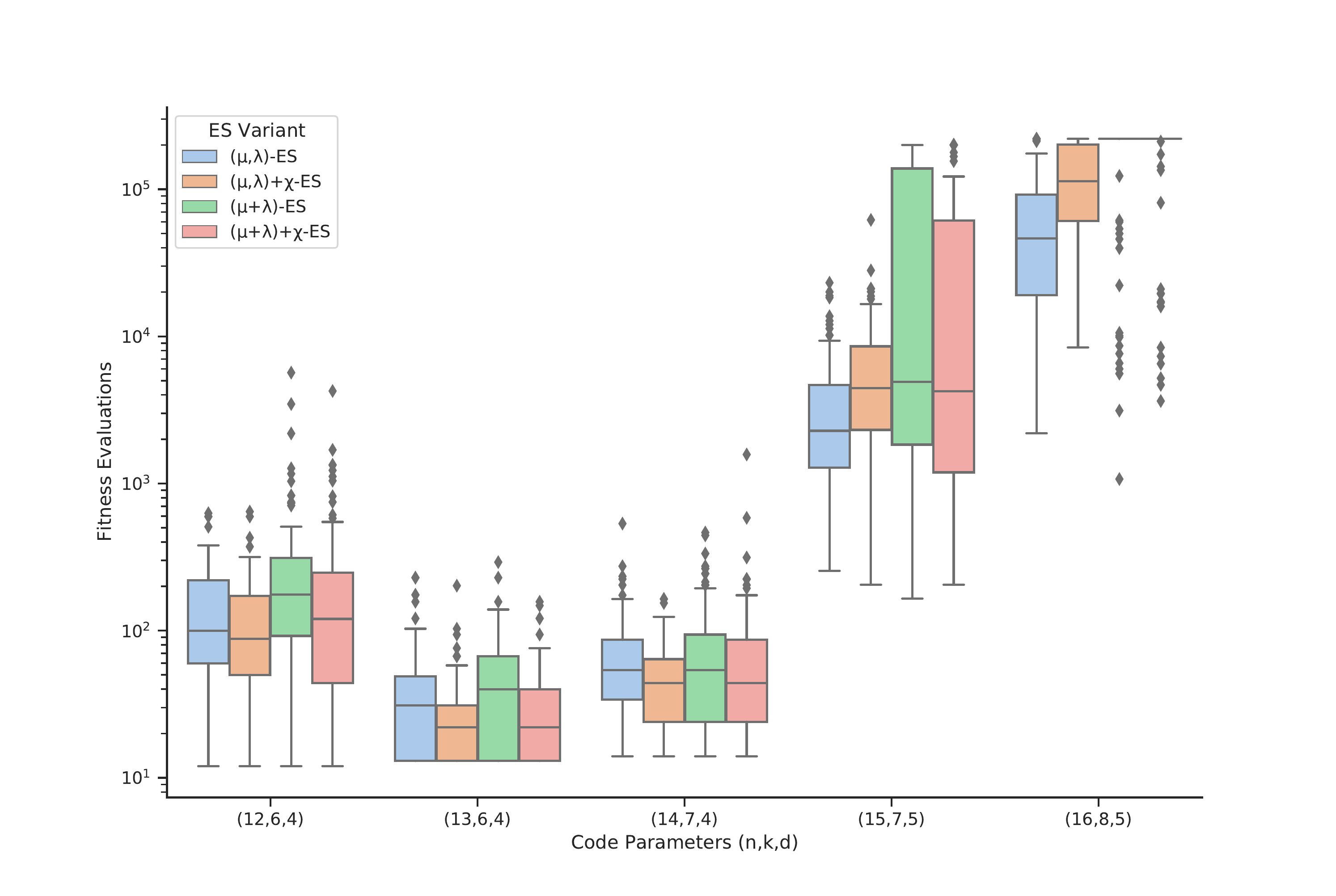}
	\caption{Fitness evaluation distributions for the four considered ES variants.}
	\label{fig:fiteval}
\end{figure}

To investigate more in detail the effects of the replacement mechanism and the crossover operator, we plotted the distributions of the number of fitness evaluations in Figure~\ref{fig:fiteval}.
In general, one can see that the number of fitness evaluations necessary to converge to an optimal solution is not directly correlated with the length of the code, and consequently with the size of the search space. As a matter of fact, the median number of evaluations of $(12,6,4)$ is always higher than that required for $(13,6,4)$ and $(14,7,4)$. Indeed, the most evident correlation is with the minimum distance, since for the two largest instances with $d=5$ the number of fitness evaluations is significantly higher. This is reasonable, since as reported in Table~\ref{tab:param}, the optimal fitness values for $d=5$ are consistently greater than for $d=4$.

As expected, for the larger instances $(15,7,5)$ and $(16,8,5)$, a clear difference emerges between the two replacement mechanisms, as already indicated by the success rates. The $(\mu,\lambda)$ variants converge to an optimal solution more quickly than the $(\mu+\lambda$) ones. On the other hand, up to $(14,7,4)$ it is not possible to distinguish the performances of the four evolutionary strategies by just looking at the respective boxplots. For this reason, we used the Mann-Whitney-Wilcoxon statistical test to compare two ES variants, with the alternative hypothesis that the corresponding distributions are not equal, and setting the significance value to $\alpha = 0.05$. The obtained $p$-values show that the $(\mu,\lambda)$-ES variants give an advantage over the $(\mu+\lambda)$-ES without crossover for $(12,6,4)$, while for $(13,6,4)$, only $(\mu,\lambda)+\chi$-ES is significantly better than $(\mu+\lambda)$-ES ($p=0.007$). For $(14,7,4)$, there is no significant difference between any two combinations of ES. Another interesting insight from the statistical test concerns the effect of the crossover operator. While for the instances up to $(14,7,4)$ there is no significant difference whether the ES is augmented with crossover or not (with the exception of $(13,6,4)$ where $(\mu+\lambda)+\chi$ is better than its counterpart without crossover, $p=0.031$), the situation is different with $(15,7,5)$ and $(16,8,5)$ for the $(\mu,\lambda)$ variants. In these cases, using crossover actually worsens the convergence speed of the ES algorithm. This is somewhat surprising, as one would expect that crossover allows to exploit the local search space more efficiently. Overall, our results show that the simplest $(\mu,\lambda)$-ES variants without crossover is the best performing one over this optimization problem.

\subsection{Solutions Diversity}
\label{subsec:sol-div}
\begin{figure}[t]
	\centering
	\includegraphics[width=0.9\textwidth]{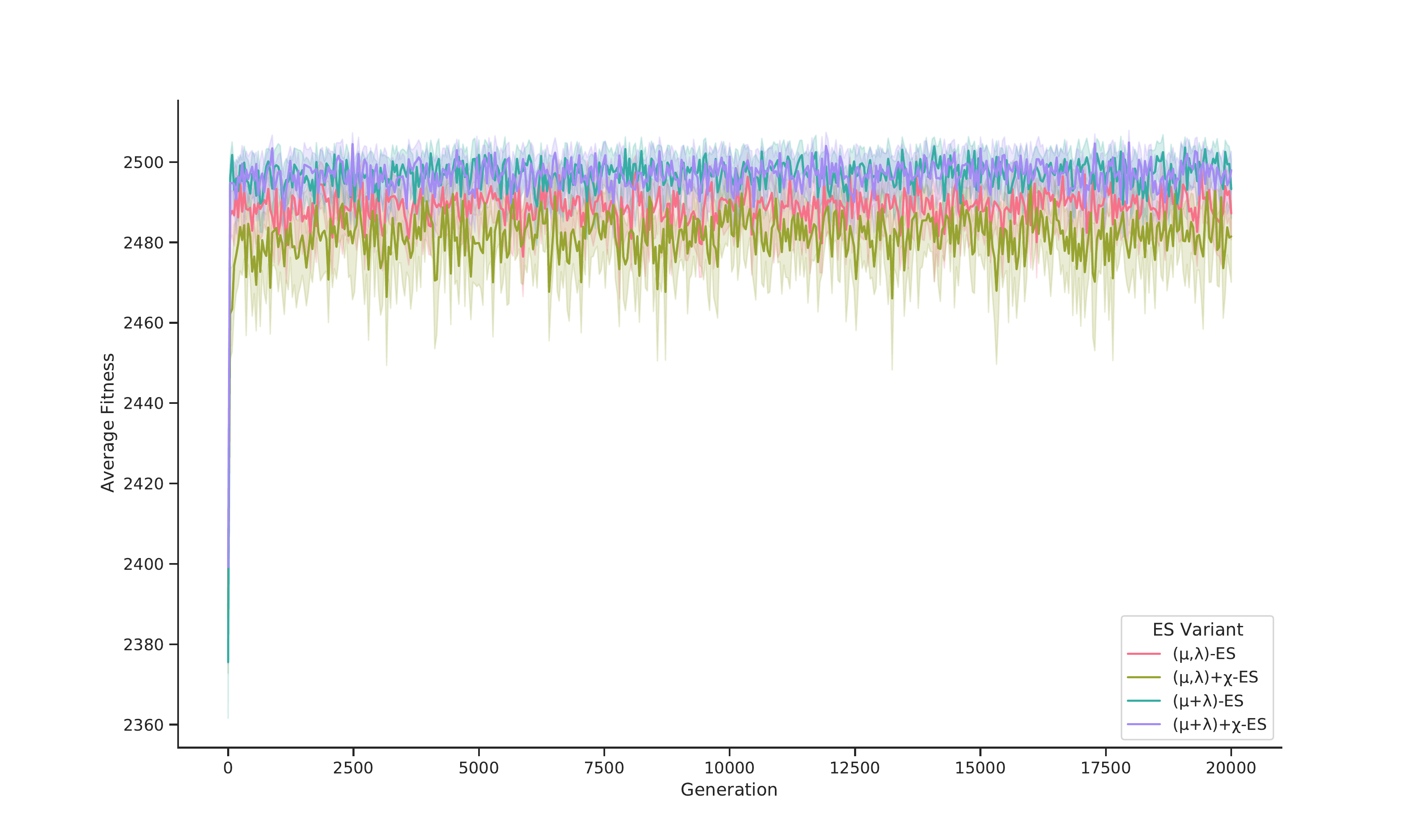}
	\caption{Average population fitness for $(16, 8, 5)$.}
	\label{fig:avgfit}
\end{figure}

\begin{figure}[t]
	\centering
	\includegraphics[width=0.9\textwidth]{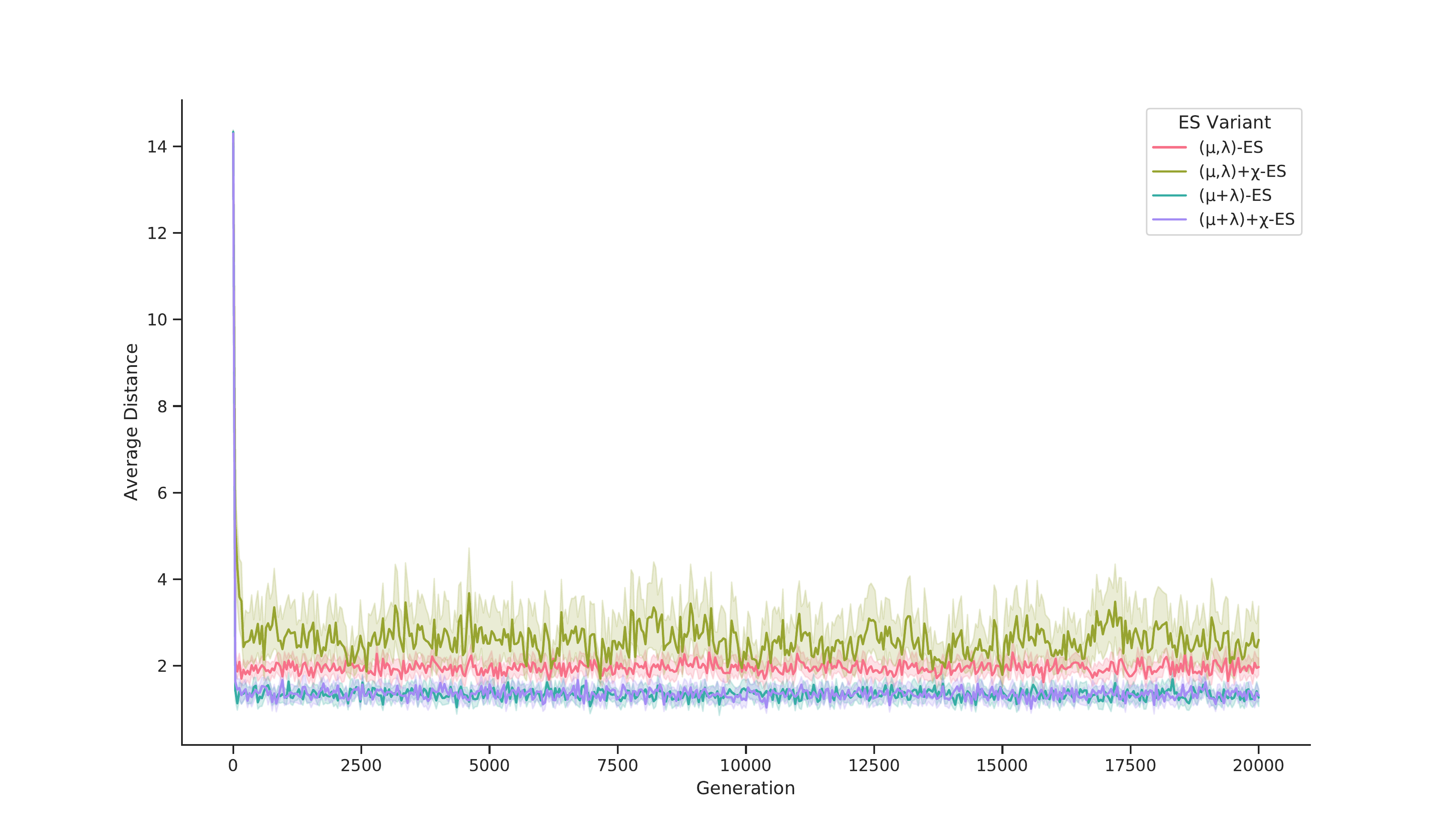}
	\caption{Average pairwise distance distributions for $(16, 8, 5)$.}
	\label{fig:avgdist}
\end{figure}

To analyze more deeply the influence of the replacement mechanism and the crossover operator on the performances of the ES algorithm, we ran again the experiments for $30$ independent repetitions on the $(16, 8, 5)$ instance, where the effects are more evident. We set the stopping criterion to $20\,000$ generations, independently of the fact that an optimal solution might be found before. In each run, we recorded every $40$ generations the average fitness of the population and the average pairwise distance between individuals, using Eq.~\eqref{eq:dist-sub}. The corresponding lineplots are displayed in Figures~\ref{fig:avgfit} and~\ref{fig:avgdist}, respectively.

It is possible to observe that the behavior of the population stabilizes almost immediately for all four ES variants. In particular, after the random initialization of the population where the fitness is relatively low and the solutions are substantially different from each other, the situation is immediately reversed in a few generations. Random perturbations continue to happen over the two measured quantities (likely due to the effect of mutation, which is used in all variants), but no huge deviation occur throughout the rest of the optimization process. In general, the population of the ES quickly converge to a highly fit area of the search space and with low diversity. This phenomenon is, however, more evident for the two $(\mu+\lambda)$ variants, which achieve the highest average fitness in the population and the lowest pairwise distance among individuals. The $(\mu, \lambda)$ variant combined with crossover is instead characterized by a slightly larger distance and lower fitness in the population, but is still very close to the $(\mu+\lambda)$ variants. On the other hand, the simple $(\mu,\lambda)$-ES is the combination reaching both the highest distance and the lowest average fitness, which is consistent with our earlier observation that this variant is the best performing one. In particular, having a higher diversity might hamper the average fitness in the population, but at the same time can help the population to escape local optima. The low diversity observed in the $(\mu+\lambda)$ variants indicates that the convex hull defined by the population under the distance of Eq.~\eqref{eq:dist-sub} shrinks very quickly, and does not grow anymore throughout the evolutionary process. Thus, if this convex hull represents a highly fit area of the search space, which however does not contain a global optimum, chances are that the population will remain stuck in that area. Likely, this effect is further strengthened by the use of crossover.  

As a final analysis, we investigated the diversity of the optimal codes produced by the four ES variants in terms of code isomorphism. Two codes $C,D \subseteq \F_2^n$ are called isomorphic if there exists a sequence of permutations on the coordinates of the codewords and on the symbols set that transforms $C$ into $D$~\cite{HuffmanP10}. We used the computer algebra system MAGMA since it has two built-in functions useful for our purpose: the function \texttt{IsIsomorphic} takes as input the generator matrices of two $(n,k,d)$ linear codes and checks whether they are equivalent up to isomorphism or not. The function \texttt{BKLC}, instead, return the generator matrix of the best known linear code for a specific combination of length $n$ and dimension $k$. In particular, all such codes reach the bound on the minimum distance reported in Grassl's table~\cite{Grassl07}, which we used as a reference to select the problem instances for our experiments. Therefore, we first used these two functions to check whether the codes produced by our ES variants are isomorphic to the best known linear codes. Further, we compared the codes obtained by the ES algorithm among themselves, to check how many isomorphism classes they belong to. Table~\ref{tab:equiv} summarizes this analysis by reporting the number of codes that are not isomorphic to the BKLC and the number of isomorphism classes, for each combination of problem instance $(n,k,d)$ and ES variant.

\begin{table}[t]
	\centering
    \scriptsize
	\begin{tabular}{ccccccccc}
		\hline
		$(n,k,d)$ & \multicolumn{2}{c}{$(\mu,\lambda)$-ES} & \multicolumn{2}{c}{$(\mu,\lambda)$+$\chi$-ES} & \multicolumn{2}{c}{$(\mu+\lambda)$-ES} & \multicolumn{2}{c}{$(\mu+\lambda)$+$\chi$-ES} \\
		\hline
		& \#non-iso & \#eq & \#non-iso & \#eq & \#non-iso & \#eq & \#non-iso & \#eq \\
		\hline
		$(12,6,4)$ & 100 & 23 & 100 & 22 & 100 & 22 & 100 & 22 \\
		$(13,6,4)$ & 100 & 85 & 100 & 81 & 100 & 78 & 100 & 79 \\
		$(14,7,4)$ & 100 & 89 & 100 & 94 & 100 & 95 & 100 & 93 \\
		$(15,7,5)$ &  72 &  5 &  63 &  6 &  51 &  5 &  44 &  5 \\
		$(16,8,5)$ &   0 &  1 &   0 &  1 &   0 &  1 &   0 &  1 \\
		
		\hline
	\end{tabular}
	\caption{Number of non-isomorphic codes to the BKLC (\#non-iso) and equivalence classes (\#eq) found by the four considered ES variants.}
	\label{tab:equiv}
\end{table}
The first remarkable finding that can be drawn from the table is that all four ES variants always discover codes that are inequivalent to the BKLC for the smaller instances with minimum distance $d=4$. Moreover, such codes belong to a high number of equivalence classes, so they are also quite diverse among themselves. From this point of view, there is also no particular difference between different ES variants. For $(15, 7, 5)$ one can remark a lower diversity since more codes turn out to be equivalent to the BKLC. Moreover, there is a noticeable difference between the $(\mu,\lambda)$ and the $(\mu+\lambda)$ variants, with the former scoring a higher number of codes inequivalent to the BKLC than the latter. Further, in general, the number of isomorphism classes drops substantially, with only 5 or 6 classes grouping all evolved codes. This phenomenon is even more extreme for the $(16, 8, 5)$ instance: in this case, all discovered codes are equivalent to the BKLC provided by MAGMA, and thus they all belong to the same equivalence class. This fact is independent of the underlying ES variant.

\section{Conclusions and Future Work}
\label{sec:outro}

To conclude, we summarize our experimental findings and discuss their relevance concerning the design of binary linear codes using evolutionary algorithms:
\begin{compactitem}
\item The proposed ES algorithm easily converges to an optimal solution for the smaller problem instances of $(12, 6, 4)$, $(13, 6, 4)$, and $(14, 7, 4)$, with no significant differences among the four tested variants. On the other hand, there is a huge increase in the difficulty of the problem for the larger instances of $(15, 7, 5)$ and $(16, 8, 5)$, although the simple $(\mu,\lambda)$-ES variant is able to maintain a very high success rate.
\item Contrary to our initial expectation, the crossover operator that we augmented our ES with either does not make any difference on the performances of the algorithm, or it even deteriorates them over the harder instances. We speculate that this is due to the small variability offered by the crossover, since it is based on the direct selection of the vectors from the parents, rather than on the vectors spanned by their generator matrices.
\item The optimal codes obtained by the ES are quite interesting from a theoretical point of view, as most of them for small instances are not equivalent to the best-known linear code produced by MAGMA, and moreover they belong to a high number of isomorphism classes. The fact that all codes instead turn out to be equivalent to the BKLC for $(16,8,5)$ is curious, and we hypothesize that this is related to the specific structure of the search space for this instance, where the global optima might be very sparse.
\end{compactitem}

Overall, our results suggest that ES represent an interesting tool to discover potentially new linear codes, and prompt us to multiple ideas for future research. One obvious direction is to apply the ES algorithm over larger instances. However, the computation of the fitness function could become a significant bottleneck in this case. Indeed, our Java implementation of the ES algorithm takes around 20 minutes to perform $20\,000$ generations on a Linux machine with an AMD Ryzen 7 processor, running at 3.6 GHz. Therefore, it makes sense to explore also with other fitness functions, maybe without relying on the characterization through the ANF of the indicator function. A second interesting direction for future research concerns the study of the variation operators proposed in this paper, especially with respect to their topological properties. In particular, we believe that both operators can be proved to be geometric in the sense introduced by Moraglio and Poli~\cite{MoraglioP05}. This might in turn give us some insights related to the structure of the Grassmannian metric space under the distance defined in Eq.~\eqref{eq:dist-sub}, and thus help us in designing better crossover operators for this problem. One idea, for instance, could be to follow an approach similar to those adopted for fixed-weight binary strings in~\cite{ManzoniMT20}.

\bibliographystyle{abbrv}
\bibliography{bibliography}

\end{document}